\documentclass[
10pt, 
a4paper, 
oneside, 
headinclude,footinclude, 
BCOR5mm, 
]{scrartcl}

\usepackage[
nochapters, 
beramono, 
eulermath,
pdfspacing, 
dottedtoc 
]{classicthesis} 

\usepackage{arsclassica} 

\usepackage[T1]{fontenc} 

\usepackage[utf8]{inputenc} 

\usepackage{graphicx} 
\graphicspath{{Figures/}} 

\usepackage{enumitem} 

\usepackage{lipsum} 

\usepackage{subfig} 

\usepackage{amsmath,amssymb,amsthm} 

\usepackage{varioref} 

\theoremstyle{definition} 

\theoremstyle{plain} 

\theoremstyle{remark} 

\hypersetup{
colorlinks=true, breaklinks=true, bookmarks=true,bookmarksnumbered,
urlcolor=webbrown, linkcolor=RoyalBlue, citecolor=webgreen, 
pdftitle={}, 
pdfauthor={\textcopyright}, 
pdfsubject={}, 
pdfkeywords={}, 
pdfcreator={pdfLaTeX}, 
pdfproducer={LaTeX with hyperref and ClassicThesis} 
} 

\usepackage[numbers]{natbib}
\usepackage[capitalize,noabbrev]{cleveref}
\usepackage{algorithm}
\usepackage{algorithmic}
\usepackage{multirow}
\usepackage{colortbl}
\usepackage{listings}

\newcommand{\ours}{\textsc{Ours}}

\title{\Large\spacedallcaps{Pretext Training Algorithms for Event Sequence Data}} 

\author{\spacedlowsmallcaps{John Smith* \& James Smith\textsuperscript{1}}} 
\author{\spacedlowsmallcaps{Yimu Wang \quad He Zhao \quad Ruizhi Deng} \\ \spacedlowsmallcaps{Frederick Tung \quad Greg Mori} \\ \spacedlowsmallcaps{Borealis AI} \\ \normalsize{\texttt{firstname.lastname@borealisai.com}}}

\date{} 

\begin{document}


\renewcommand{\sectionmark}[1]{\markright{\spacedlowsmallcaps{#1}}} 
\lehead{\mbox{\llap{\small\thepage\kern1em\color{halfgray} \vline}\color{halfgray}\hspace{0.5em}\rightmark\hfil}} 

\pagestyle{scrheadings} 


\maketitle 

\setcounter{tocdepth}{2} 

\tableofcontents 




\section*{Abstract} 

Pretext training followed by task-specific fine-tuning has been a successful approach in vision and language domains. This paper proposes a self-supervised pretext training framework tailored to event sequence data. We introduce a novel alignment verification task that is specialized to event sequences, building on good practices in masked reconstruction and contrastive learning.  Our pretext tasks unlock foundational representations that are generalizable across different downstream tasks, including next-event prediction for temporal point process models, event sequence classification, and missing event interpolation. Experiments on popular public benchmarks demonstrate the potential of the proposed method across different tasks and data domains.





\newpage 


\section{Introduction}

Self-supervised learning has emerged as an effective machine learning paradigm in domains including computer vision~\cite{mae}, natural language processing~\cite{gpt3}, and time series modeling~\cite{ts2vec}. A common two-step strategy consists of performing large-scale unsupervised training followed by task-specific fine-tuning. During unsupervised training, a pretext training task, such as the masked reconstruction~\cite{mae} of pixels or words, is used to extract general representations from unlabelled data. The supervised fine-tuning step then adapts the model to downstream tasks of interest (alternatives to fine-tuning, such as prompt-tuning \cite{llmtime2} in large language models, are also possible).

Event sequence data are frequently encountered in areas including commerce, science, and healthcare: for example, sequences of online purchases, banking transactions, earthquakes, disease outbreaks, and hospital patients' medical measurements. These sequences comprise discrete events that are irregularly spaced in time. They are often modelled as temporal point processes, realizations of which are sequences $\{(t_1, m_1), ..., (t_N, m_N)\}$, where $N$ is the number of events (a random variable), $\{t_{i}\}$ are event arrival times and $\{m_{i}\}$ are event marks (e.g., event type). When event data are plentiful, large-scale pretext training can enable the learning of general representations without labels toward the building of foundational capabilities. For example, pretext training on banking transactions could help produce representations that are useful for detecting financial crime, for which labels are scarce; it could also provide foundational capabilities to support a range of client-oriented features such as personal budgeting or product recommendation.

\begin{figure}[t]
\centering
\includegraphics[width=.9\linewidth]{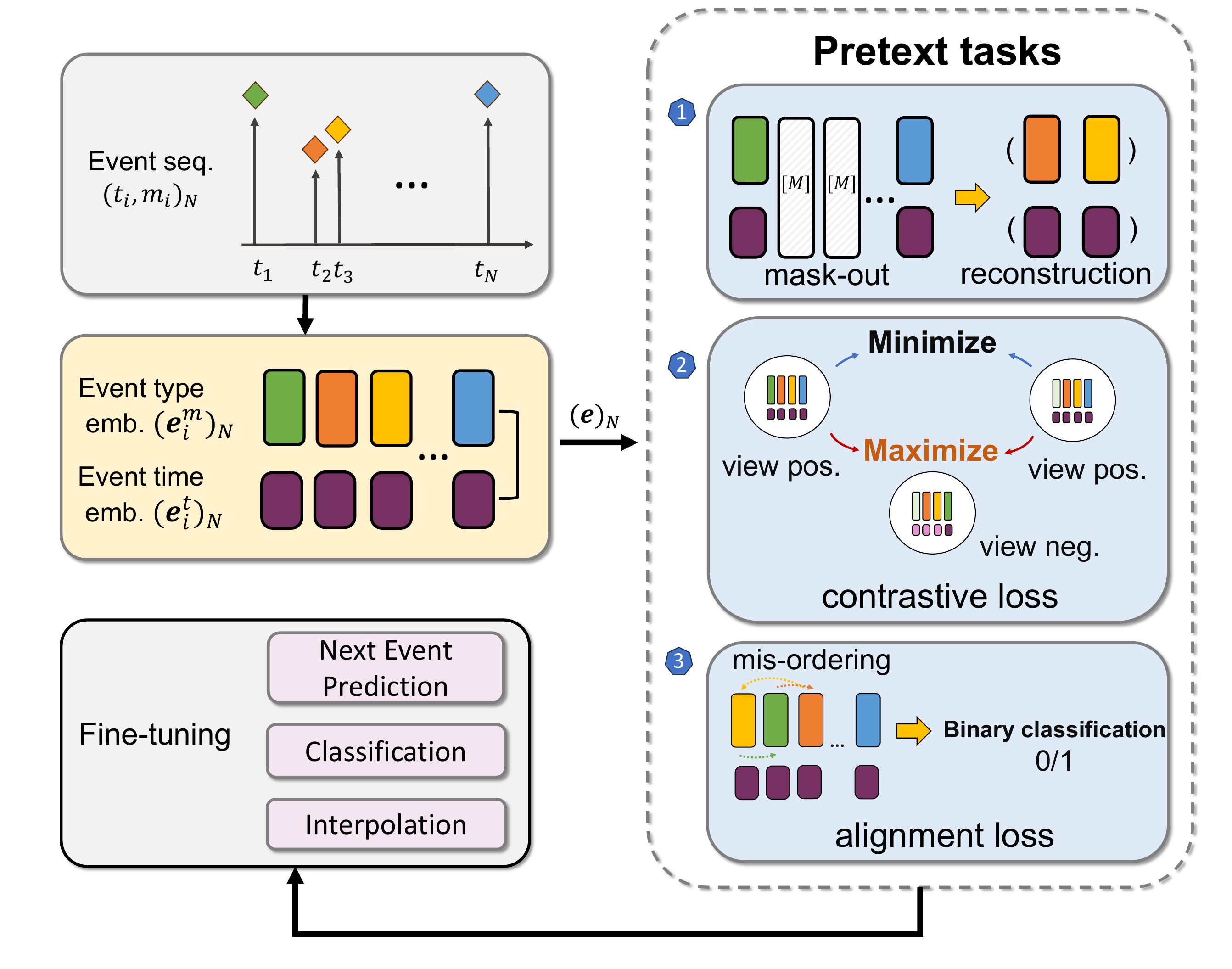}
   \caption{Overview of the pretext training and fine-tuning framework for event sequences. In the first stage, event sequence embeddings are input to self-supervised pretext tasks based on masked reconstruction, contrastive learning, and alignment verification, to learn general representations. The second stage performs task-specific fine-tuning to support downstream tasks such as next-event prediction, sequence classification, and missing event interpolation.
   }
\label{fig:teaser}
\end{figure}

In this paper, we seek the answer for: \textit{what are good pretext tasks for event sequence data?} We focus on three pretext tasks: masked reconstruction, contrastive learning, and misalignment detection. While these three pretext tasks may be used independently, we show that they are in fact complementary, and the best empirical outcomes are obtained by using them jointly. First, we build on masked reconstruction and contrastive learning, inspired by their success in computer vision and natural language processing. Next, we introduce a new pretext task that is specialized to event sequence data. Exploiting the coupling of event time and type, we propose a novel alignment verification task. To learn representations that capture the coupling relationships, we train a binary classifier to recognize misaligned event sequences, which can be easily generated by mixing up or disordering the original sequence. Finally, we show that simply appending task heads (e.g., multilayer perceptrons) and fine-tuning the model can yield strong performance on downstream tasks.

\section{Related work}

\paragraph{Pre-training in machine learning} 
The goal of pre-training is to learn generalizable and transferable representations that can be adapted to diverse downstream tasks, for example by fine-tuning.
In language modeling, BERT~\cite{bert} revolutionized the research landscape by presenting two pre-training tasks: masked language model and next sentence prediction. These led to powerful models that can excel across various tasks, including question answering, sentiment analysis, and named entity recognition.
Similarly, GPTs~\cite{gpt3,gpt2} found that simply using next-token prediction as the pre-training objective over a larger scale of data and model capacity can yield significant improvements.
In computer vision, supervised ImageNet~\cite{imagenet} pre-training had a wide-ranging impact on object detection~\cite{fasterrcnn}, segmentation~\cite{swintransformer} and cross-modal retrieval~\cite{clip}. More recently, unsupervised pre-training has been used to learn robust and useful representations from image data without labels~\cite{mae}. 
In time series analysis, recent works following approaches similar to the above have made great progress in regular time series~\cite{unsupervisedtime1, tnc, ts2vec, unsupervisedtime3} and irregular time series~\cite{primenet} on diverse tasks, including classification, interpolation, and forecasting. However, pre-training methods are not prevalent for event sequence data.
Common across these approaches is the establishment of simple and effective pretext tasks for model pre-training. 

\paragraph{Pretext tasks} A widely used pretext task in many research domains is \textbf{maskout}-\textbf{reconstruction} \cite{mae,bert}. These efforts randomly remove part of the data (\textit{e.g.}, pixels, words, or time series values) and train the model to fill in the masked part through a high-level understanding of the neighboring context.
Our approach adopts a similar idea but tailors the mask token sampling strategy specific to event sequence data. 
\textbf{Contrastive learning} is another popular pretext task where the goal is to bring different views of the same data together while
pushing views of different data apart~\cite{simclr}. To construct different views, a variety of augmentation schemes has been explored; for reviews, see~\cite{goodviewcl, goodviewcltime}. However, it remains unclear if they can apply to event sequence data. In this work, we leverage simple techniques to construct such views: (i) sub-sequence sampling, (ii) adding noise to embeddings, and (iii) randomly masking out data. Moreover, \textbf{learning via alignment} approaches have seen significant interest in cross-modality learning research.
The approach 
defines a binary classification problem
of predicting misaligned vs. aligned data (\textit{e.g.}, shuffled video clips) as a pretext task for pre-training~\cite{align1, univl, align2, align3}. We apply their insight on the use of alignment loss to pre-train our event sequence model, with a particular focus on verifying the correct coupling of event type and time.

\paragraph{Temporal point processes} 
A temporal point process (TPP)~\cite{tppintro} is a mathematical tool for modeling sequences of asynchronous discrete events. 
Depending on their approaches to modeling the distribution of events in time, TPPs can be broadly categorized into intensity-based~\cite{hawkes2, hawkes1} and intensity-free models~\cite{intensityfreetpp}. 
The former approach parameterizes an intensity function of time to determine the probability of event occurrence. 
In contrast, the latter approach directly models the distribution of time intervals between event occurrences.
Models for sequential data such as recurrent neural networks have been an important building block of neural TPP models~\cite{mehrasa2019variational, tpplstm}. More recently, TPP efforts focus on using transformer blocks~\cite{transformer} to build attentive TPPs for better performance~\cite{anhp} as well as leveraging more advanced learning formulations, such as meta-learning~\cite {metatpp}.
Recent surveys~\cite{tppreview, easytpp} describe a variety of TPP models proposed over the past decade.


\section{Technical approach}

We develop novel pretext tasks that are effective for event sequences. This section presents (i) a maskout and reconstruction task that is applied on both event type and time, (ii) a contrastive learning task that uses simple and effective augmentations to generate positive and negative views of data, and (iii) a novel alignment verification task that enforces the models to learn the intrinsic consistency between feature dimensions. All pretext tasks are independent of the architecture of the model and downstream tasks, and do not require any labeled data. Interestingly, we find that the three tasks are in fact complementary: the best empirical results are achieved when they are used together.

\subsection{Preliminaries}

Let $X = \{(t_1, m_1), ..., (t_N, m_N)\}, \, 0 < t_1 < ... < t_N $ denote an event sequence where $N$ is the total number of events, $\{t_{i}\}$ are event arrival times and $\{m_{i}\}$ are event types (e.g., categorical variables from domain with size $K$).
The time intervals of such a sequence are denoted $(\tau_1, ..., \tau_{N-1})$, where $\tau_i = t_{i+1}-t_i$.

We introduce pretext tasks to pre-train a backbone network for representation learning that can be subsequently fine-tuned for various downstream tasks, such as
modelling the probability distribution of the next event type and time.
In our implementation, we adopt the transformer architecture in \citet{anhp} for representation learning. 
Next, we briefly describe how event information is encoded under this setting.

\paragraph{Event time embeddings} Following recent advances, we use positional encodings~\cite{anhp, transformer} to transform each event arrival time into a vector, $\textbf{e}^t \in \mathcal{R}^{D_{time}}$.

\paragraph{Event type embeddings} We simply encode categorical event types into a high dimensional vector, $\textbf{e}^m \in \mathcal{R}^{D_{type}}$, with a learnable embedding layer, similar to~\citet{rnntpp}. 

\paragraph{} The input representation of an event is the concatenation of $\textbf{e} = [\textbf{e}^m, \textbf{e}^t]$. 

\subsection{Masked reconstruction}\label{sec:mae}

Event data can be viewed as sequences of tokens similar to natural language sentences. This leads us to conjecture that it is valuable to capture global or local contextual information. To this end, the first pretext task we present is to reconstruct the masked event data.  

\paragraph{Density preserving masking} 
Masked reconstruction is one of the most popular self-supervised learning approaches in vision and language, and has recently extended to the time series domain with masking of continuous fragments~\cite{tnc,ts2vec,tst}. We adopt a density-preserving masking strategy~\cite{primenet} motivated by the non-uniform nature of event arrival times. This approach randomly samples intervals with constant time duration and masks all the events in the intervals.
The underlying hypothesis is that time intervals with dense events contain more contextual information than sparse intervals, making the reconstruction of events easier in dense intervals. Therefore, an ideal masking strategy would preserve the original density of event arrival times, \textit{i.e.}, masking more events when event arrivals become more frequent. The constant duration (density-preserving) masking fits this need.
Following common practice~\cite{bert,mae}, the masked events are replaced by a learnable $\mathtt{[MASK]}$ token.

\paragraph{Reconstruction} 
We feed the sequence consisting of (i) visible (non-masked) event embeddings and (ii) mask tokens to our encoder, and extract embeddings corresponding to the masked out locations.
The embeddings are decoded to reconstruct the time and type embeddings of the masked events and we use mean squared error (MSE) to supervise the training of this pretext task.
Denote the set of masked event indices by $M$.
The masked reconstruction loss is
\begin{equation}
    \mathcal{L}_{rec} = \frac{1}{|M|}\sum_{i\in M}\|\mathbf{e}^t_i -\tilde{\mathbf{e}}^t_i\|_2^2 + \|\mathbf{e}^m_i - \tilde{\mathbf{e}}^m_i\|_2^{2},
\end{equation}
where $\tilde{\mathbf{e}}^t_i$ and $\tilde{\mathbf{e}}^m_i$ denote the reconstructed event time and type embeddings of $(t_i, m_i)$ respectively and $|M|$ is the number of masked events. 

\subsection{Contrastive learning}\label{sec:cl}

We use contrastive learning to pull similar event sequences closer together in representation space and push dissimilar event sequences further apart.

\paragraph{Data augmentation} 
The success of contrastive learning relies on augmentation methods to create multiple views of data. 
We investigate three simple and effective data augmentation methods as follows.

\paragraph{View 1: Subsequence} We randomly extract subsequences from the original data and treat them as novel views, which is a strategy that has been found to be effective in other fields~\cite{primenet, ts2vec}. This method is conceptually intuitive: Subsequences of event data can be seen as the counterparts of patches or croppings in vision research, representing a local division of original data.
Our approach creates subsequences by taking events in a continuous time period in the original sequences.

\paragraph{View 2: Masked events}
The second method further exploits the masked data (in a continuous period) proposed in \cref{sec:mae} to create novel views. Previous masked autoencoder research~\cite{mae} and our first pretext task both show that the model can learn to reconstruct masked data from contextual information. Therefore, we assume that masked data should contain similar semantic information
as the original data, and contrastive learning based on masked data can further enhance the separation between similar and non-similar sequences. 
We use the learnable special token $\mathtt{[MASK]}$ as in \cref{sec:mae}
 to mask the selected events.
 
\paragraph{View 3: Noisy data} We add multi-scale Gaussian noise to the event sequence embeddings and take them as novel views. 
Compared to \textbf{View} \textbf{1} and \textbf{2}, 
noise injection is a type of augmentation closer to real-world data corruption. 
A good representation should also be robust to noise corruption~\cite{denoisingae}. Therefore, we conjecture that noisy versions of event sequences 
can be effective in creating multi-view data for contrastive learning. Specifically, we take the following steps to inject noise into data: (i) Uniformly sample the scale of noise, $\sigma \in [0, 1]$, and (ii) sample Gaussian noise from $\mathcal{N}(0,\,\sigma)$. The sampled noise is added to the embeddings.

\begin{algorithm}[t]
    \caption{The pseudo-code for generating different views of data in contrastive learning.}\label{alg. contrastive learning}
    \begin{algorithmic}
        \STATE \textbf{View 1: Subsequence} (event sequence $X$, ratio $r$):
        \STATE[1.] Get the subsequence length $l\leftarrow r(t_N - t_1)$;
        \STATE[2.] Randomly sample a start time $t_{start} \leftarrow uniform([t_1, t_N - l])$;
        \STATE[3.] Return the subsequence $X_{view1}$ which only contains the events that fall in the sampled continuous time period $[t_{start}, t_{start} + l)$.
        \STATE 
        \STATE \textbf{View 2: Masked events} (event sequence $X$, ratio $r$):
        \STATE[1.] Get the subsequence length $l\leftarrow r(t_N - t_1)$;
        \STATE[2.] Randomly sample a start time $t_{start} \leftarrow uniform(t_1, t_N - l)$;
        \STATE[3.] Return the masked sequence $X_{view2}$ which replaces the events that fall in the sampled continuous time period $[t_{start}, t_{start} + l)$ by $\mathtt{[MASK]}$.
        \STATE 
        \STATE \textbf{View 3: Noisy data} (event sequence $X$):
        \STATE[1.] Get the embedding $\mathbf{z}$ of the event sequence $X$ at the $\mathtt{[EOS]}$ token by forwarding;
        \STATE[2.] Uniformly sample the scale $\sigma \in [0, 1]$;
        \STATE[3.] Sampling Gaussian noise $\mathbf{n} \sim \mathcal{N}(\textbf{0},\,\sigma)$;
        \STATE[4.] Return the noisy embedding $\tilde{\mathbf{z}} \leftarrow \mathbf{z} + \mathbf{n}$.
    \end{algorithmic}
\end{algorithm}

\paragraph{} The generation procedure of the three views is summarized in \cref{alg. contrastive learning}. 
To compute the contrastive loss, we need to represent the entire sequence with one vector. To this end, we append a special $\mathtt{[EOS]}$ token to the end of event sequences and use the output embeddings corresponding to $\mathtt{[EOS]}$, \textit{i.e.}, $\mathbf{z}$, as the aggregated representation. We use all three augmentation methods for each data in the mini-batch and adapt the normalized temperature-scaled cross-entropy loss (NT-Xent)~\cite{simclr} to fit our setting.
Given a mini batch of training data of size B, we use $S_i$ to denote the set consisting of the embedding of the $i$th sample and the embeddings of all its three augmented views, and $S = \bigcup_{i=1}^N S_i$ to denote the union of all $S_i$s. Denote $P_{S_i}^2$ as the set of all combinations of two different embeddings from $S_i$ with permutation (positive pairs), which is $\{(z, z' ) | z, z' \in S_i \text{ and } z \neq z' \}$. 
Given $k(=3)$ augmented views, the total number of positive pairs in $P_{S_i}^2$ is $k\cdot (k+1)$.
We define the contrastive loss for the $ith$ sample $l_i$ as 
\begin{equation}
    l_i = -\frac{1}{|P_{S_i}^2|}\sum_{(z, z')\in P_{S_i}^2}\log \frac{\exp(z\cdot z'/\eta)}{\sum_{z''\in S-\{z\}} \exp(z\cdot z''/\eta)}.
\end{equation} The final contrastive loss $\mathcal{L}_{cl}$ is calculated as $\mathcal{L}_{cl} = \frac{1}{B}\sum_{i=1}^B l_i$.

\subsection{Alignment verification}
We observe two important properties of event data: (i) Event types are associated with the corresponding arrival times, and (ii) preceding events can decide the subsequent events. 
For instance, the event of ``lunch break'' is often associated with noon times; and the occurrence of ``buying flight tickets'' often precedes ``booking hotel''. A well-trained model should recognize any violation of these relationships. Hence, we explore the utility of verifying such alignment as model pre-training.

More specifically, this pretext task is designed as a binary classification problem, aiming to tell apart correctly aligned event sequences from misaligned ones. 
To generate misaligned event sequences for pre-training, we devise the following methods.

\paragraph{Misalignment 1: Shuffle} 
Randomly shuffling the event types both disrupts the consistency between event types and event arrival times and breaks their correct order. We keep the event arrival times intact so that they still follow the same distribution as the original sequences and the training focuses on learning better contextual representations of event types.

\paragraph{Misalignment 2: Swap} Similarly, we can create a misalignment by mixing event types of a given sequence with event times of another sequence. This can be done by randomly picking a different event sequence in the same batch and swapping their type and time dimensions\footnote{In implementation, we pad all sequences to the same length.}.

\paragraph{Misalignment 3: Crossover} The previous two approaches focus on creating misalignment on one feature dimension, \textit{e.g.}, the event time. 
Here, we go one step further 
and adopt the crossover operator from genetic algorithms~\cite{genetic}
to generate misaligned data,
combining halves of two randomly selected sequences on both event type and time. 
We ensure monotonicity in arrival time values by crossing over the time intervals 
from the other sequence instead of the absolute arrival times $t$.

\begin{algorithm}[t]
    \caption{The pseudo-code for generating misaligned sequences in alignment verification.}\label{alg: alignment}
    \begin{algorithmic}
        \STATE \textbf{Misalignment 1}: \textbf{Shuffle} (event sequence $X$):
        \STATE[1.] Get the shuffled indices by $\tilde{ind}\leftarrow \operatorname{Shuffle}([N])$;
        \STATE[2.] Return the event sequence shuffled along the type dimension $\tilde{X} = \{(t_1, m_{\tilde{ind}_{1}}), \ldots, (t_N, m_{\tilde{ind}_{N}})\}$.
        
        \STATE 
        \STATE \textbf{Misalignment 2:} \textbf{Swap} (event sequences $X_a, X_b$):
        \STATE Return the swapped event sequences
        $\tilde{X}_a = \{(t_{[a, 1]}, m_{[b,1]}), \ldots, (t_{[a, N]}, m_{[b,N]})\}$ and 
        $\tilde{X}_b = \{(t_{[b, 1]}, m_{[a,1]}), \ldots, (t_{[b, N]}, m_{[a,N]})\}$.

        \STATE 
        \STATE \textbf{Misalignment 3:} \textbf{ Crossover} (event sequences $X_a, X_b$):
        \STATE Assuming even lengths $N_a, N_b$ for ease of presentation \\
        \STATE Return the combined event sequences \\
        $\tilde{X}_a = \{(t_{[a, 1]}, m_{[a,1]})$, $\ldots, (t_{[a, N_{a}/2]}, m_{[a,N_{a}/2]})$, 
    $ (t_{[a, N_{a}/2]} + t_{[b, N_{b}/2+1]} - t_{[b, N_{b}/2]}, m_{[b, N_{b}/2 + 1}])$, $\ldots$, $(t_{[a, N_{a}/2]} + t_{[b, N_b]} - t_{[b, N_{b}/2]}, m_{[b,N_{b}]})\}$ and 
        $\tilde{X}_b = \{(t_{[b, 1]}, m_{[b,1]})$, $\ldots, (t_{[b, N_{b}/2]}, m_{[b,N_{b}/2}])$, 
        $ (t_{[b, N_{b}/2]} + t_{[a, N_{a}/2+1]} - t_{[a, N_{a}/2]}, m_{[a, N_{a}/2 + 1}])$, $\ldots$,  $(t_{[b, N_{b}/2]} + t_{[a, N_a]} - t_{[a, N_{a}/2]} , m_{[a,N_{a}]})\}$.
    \end{algorithmic}
\end{algorithm}

\begin{figure}[thp]
\centering
\subfloat[Misalignment 1: Shuffle]
{\includegraphics[width=\textwidth]{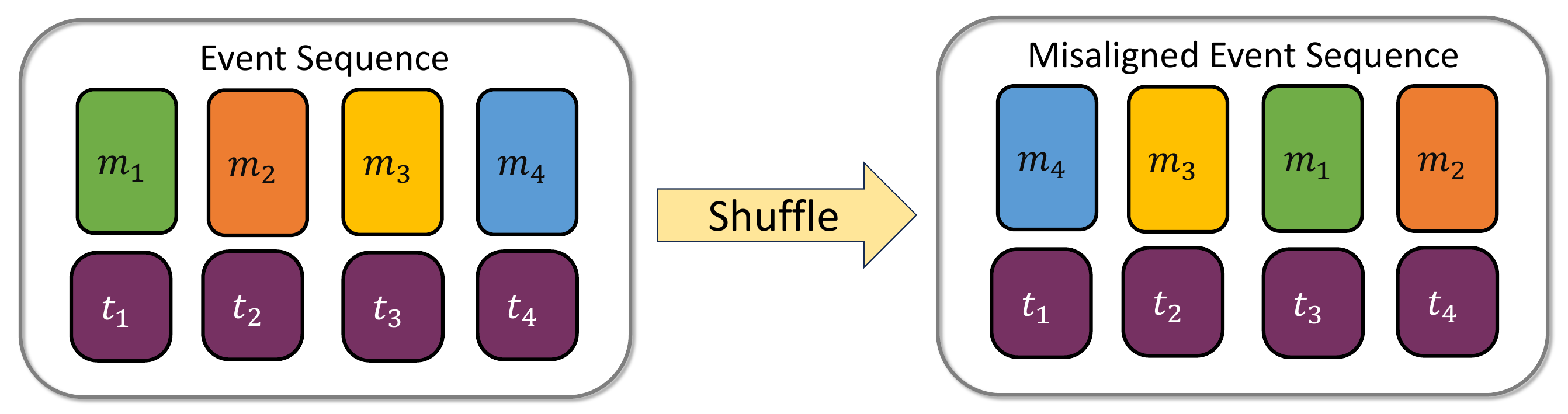}}\\
\subfloat[Misalignment 2: Swap]
{\includegraphics[width=\textwidth]{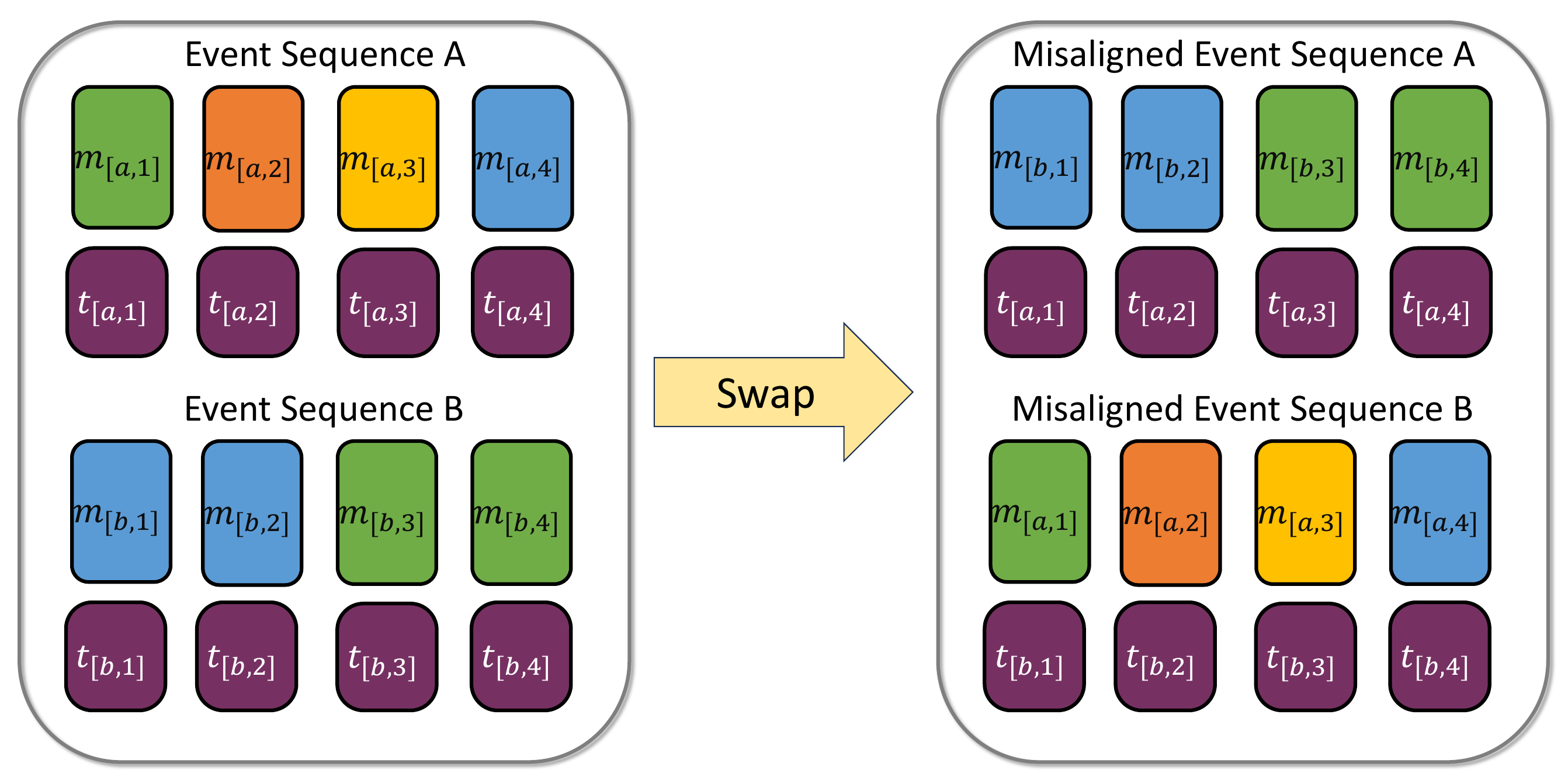}}\\
\subfloat[Misalignment 3: Crossover \label{align:crossover}]
{\includegraphics[width=\textwidth]{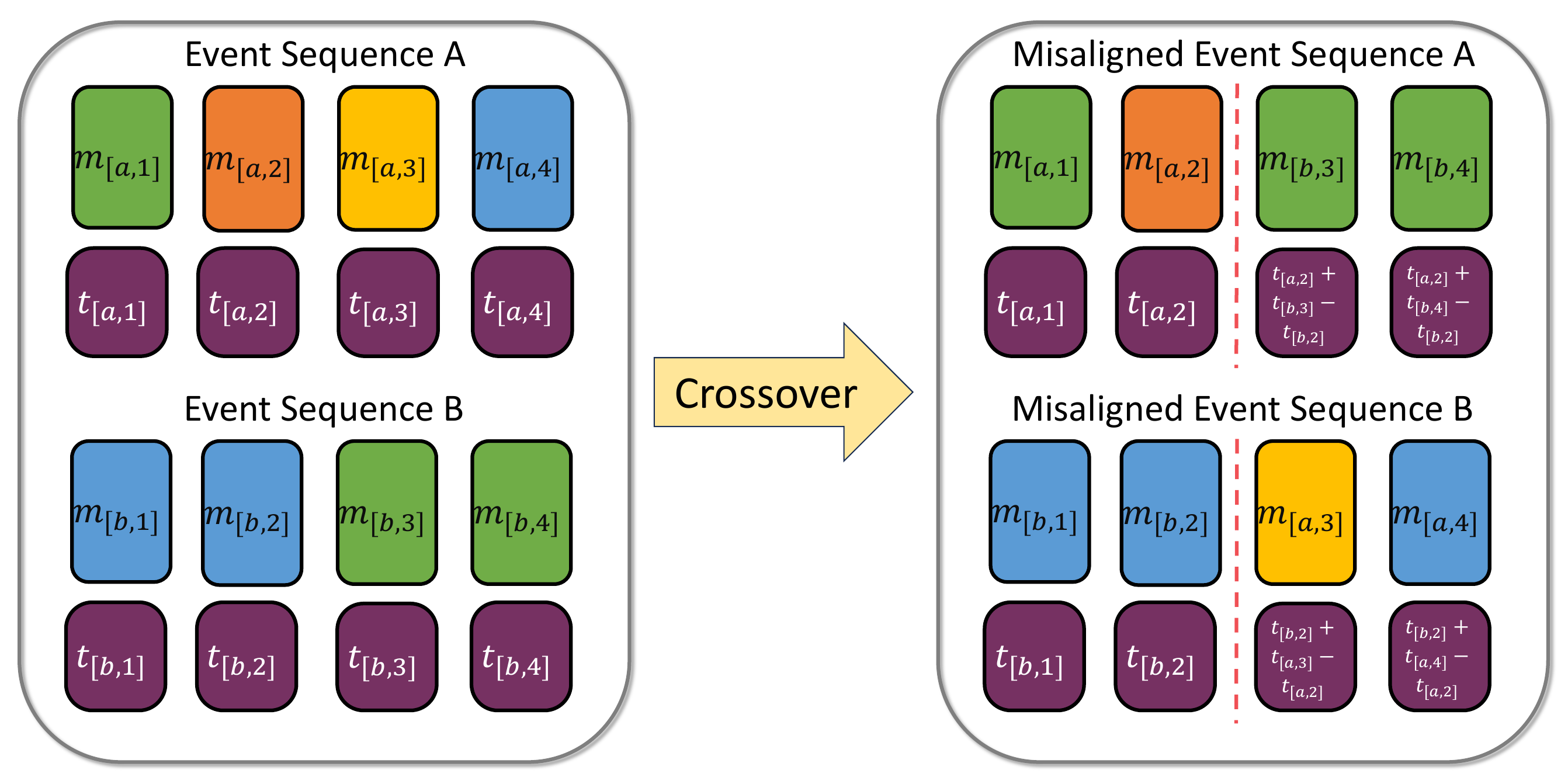}}
\caption{
Visualization of the three approaches we use to create misaligned event sequences: shuffle, swap and crossover. For illustrative purposes, we show the approaches with event sequences of length 4. The red dashed lines in Figure~\ref{align:crossover} indicate the cut-off point of crossover.
}
\label{align}
\end{figure}

\paragraph{} The three misalignment generation methods are summarized in \cref{alg: alignment} and visualized in \cref{align}.
Similar to ~\cref{sec:cl}, we use the feature output of the $\mathtt{[EOS]}$ token as input to an MLP classifier and compute the binary cross-entropy loss, $\mathcal{L}_{alignment}$.

\subsection{Learning objective}
\paragraph{Pretext training} 
Finally, we calculate the loss as the weighted sum of these pretext tasks according to
\begin{equation*}
    \mathcal{L}_{pretext} = \alpha\; \mathcal{L}_{rec} + \beta\; \mathcal{L}_{cl} + \gamma\; \mathcal{L}_{alignment},
\end{equation*}
where $\alpha, \beta,$ and $\gamma$ are the combination weights. 

\paragraph{Downstream tasks} 
To show the generality of our proposed methods, we choose three representative downstream tasks in event sequence analysis, \textit{i.e.}, temporal point process, event sequence classification, and missing events imputation. The details are presented in \cref{sec: downstream tasks}.

\section{Experiments}

In this section, we evaluate our self-supervised representations on a variety of downstream tasks, including next-event prediction for temporal point process models, event sequence classification, and missing event imputation. We compare with a broad range of baselines, including state-of-the-art temporal point process models and popular pre-trained LLMs in a zero-shot setting.

\subsection{Downstream tasks, datasets and evaluation protocol}\label{sec: downstream tasks}

We follow a standard pretext-finetuning paradigm using our pretext training methods and evaluate them on three representative downstream tasks as described below.

\paragraph{Downstream task 1: Temporal Point Process (TPP)}
This task aims at modeling the intensity
$\lambda\left(t \right)$
at a given time $t$.
During fine-tuning, we directly minimize the negative log-likelihood loss as follows,
\begin{equation}\label{eq:NLL}
\begin{aligned}
    \mathcal{L}_{finetuning} =& -\underbrace{\sum_{i\in[N]} \log \lambda\left(t_i \right)}_{\text {event log-likelihood }} +\underbrace{\int_{t_1}^{t_N} \lambda\left(t  \right) d t}_{\text {non-event log-likelihood }}. 
\end{aligned}
\end{equation}
Following~\cite{hawkes1}, we conceptualize the intensity function for each event category, $\lambda_{k}$, as $\operatorname{softplus} (\alpha_k \frac{t - t_j}{t_j} + \mathbf{w}_k^\top \mathbf{h}_t + b_k)$,
where $t_j$ is the time of the preceding event, $\mathbf{h}_t$ the hidden states and $\{\alpha_k, \mathbf{w}_*, b_*\}$ are learnable parameters. The overall intensity is aggregated as $\lambda = \sum_{k=1}^{K} \lambda_{k}$.

\paragraph{Downstream task 2: Event sequence classification}
This task aims at predicting a sequence-level label. 
During fine-tuning, we minimize the binary cross-entropy loss.

\paragraph{Downstream task 3: Missing events imputation}
This task 
aims at reconstructing missing events, and simulating partial observability in a noisy environment.
During fine-tuning, we minimize the cross-entropy loss and root-mean-square error (RMSE) for 
the missing values of event type and time.

\paragraph{Benchmark datasets}
We evaluate our proposed method on four real-world datasets, \textit{i.e.}, StackOverflow~\cite{snapnets}, Mooc~\cite{kumar2019predicting}, Reddit~\cite{kumar2018community}, and MIMIC-II~\cite{DBLP:conf/embc/LeeSVCSM11}.
Details are presented in
\cref{appendix:sec:datasets}. 

\paragraph{Baselines}
For TPP, 
we compare with several state-of-the-art temporal point process models 
reported in recent work~\cite{metatpp, anhp}.
Moreover, following advances in large language models~\cite{llmtimeseries},
we produce next event predictions with popular LLMs using a zero-shot setting. 
For the other two tasks, 
we consider a baseline: the same model as ours that is trained from scratch for the specific downstream task. 

\paragraph{Metrics}
For TPP, we report the negative log-likelihood (NLL)
and root mean squared error (RMSE) for event time prediction, and accuracy for event type prediction. 
For binary classification, we report the area under the curve (AUC).
For missing events imputation, we report accuracy and RMSE.

\subsection{Implementation details}

\paragraph{Architecture details}
We adopt the transformer encoder in \citet{anhp} as our backbone for representation learning.
This module is configured with three encoder layers and each layer is multi-head attention implemented with four heads and feature dimensions as $64$ ($D_{time} = D_{type} = 32$). 

\paragraph{Pretext training details} 
(i) For masked reconstruction, we recover the masked event type and time embeddings with two additional decoders.
We find that a 30\% masking ratio on time duration for density-preserving masking is sufficient as presented in \Cref{tab:ablation_reconstruction_ratio}.
(ii) For contrastive learning, we use the embedding at the special token $\mathtt{[EOS]}$ (see ~\cref{sec:cl}) to compute the loss with $r = 0.3$ for subsequences and masked events generation. 
(iii) For alignment verification, we use the same embedding and append a linear layer as the binary classifier. 
In pretext training, the learning rate is fixed at 1e-4, the number of epochs is $10$, and the batch size is set to $4$. 
For simplicity, we set $\alpha = \beta = \gamma = 1$. 

Once pretext training is complete, the backbone acts as the representation extractor and initial point for fine-tuning. At this stage, task-specific heads are added for each task. 

\paragraph{Downstream fine-tuning details} 
For every downstream task, we fine-tune the model and task head for $300$ epochs with a batch size of $4$ and a learning rate of 1e-4. The task heads are simple MLPs. 

For TPP,
the task head yields parameterized intensity functions in \cref{eq:NLL}. 
For event sequence classification,
the head is applied
on the embedding of $\mathtt{[EOS]}$ to predict the class probabilities.
For imputation,
the two MLP heads predict time (scalar regression) and type (classification).

\subsection{Main results}
This subsection presents our main experimental results on three downstream tasks: next-event prediction for temporal point process models, sequence-level classification, and missing events imputation.

\begin{table}[tp]
\centering
\caption{The results of TPP (NLL, RMSE, and accuracy) on Stack Overflow, MIMIC-II, Mooc, and Reddit datasets. ``$\dagger$'' refers to numbers from \citet{metatpp}. 
``THP+'' is THP with a mixture of log-normal distribution decoder produced by \citet{metatpp}.
Standard deviations are shown in subscripts.
}
\begin{tabular}{lcccc}
\toprule
Methods  & NLL $ \downarrow$              & RMSE $\downarrow$ & Acc $ \uparrow$  \\ \midrule
\multicolumn{4}{c}{Stack Overflow} \\ \midrule 
THP & $2.42_{0.00}$ & $1.78_{0.02}$ & $46.52_{0.01}$
\\
SAHP & $2.26_{0.04}$ & $1.91_{0.10}$ &  $46.87_{0.00}$
\\
Intensity Free$\dagger$ & $3.66_{0.02}$ & $3.64_{0.26}$ & $43.00_{0.01}$\\
TPP+$\dagger$ & $3.28_{0.02}$ & $1.68_{0.16}$ & $46.00_{0.00}$\\
Meta TPP$\dagger$ & $2.64_{0.02}$ & $1.15_{0.02}$ & $46.00_{0.01}$\\
ANHP & $2.16_{0.02}$ & $1.19_{0.01}$ & $47.42_{0.00}$  \\
\rowcolor[gray]{0.90} \ours & $\textbf{1.81}_{0.05}$ &        $\textbf{1.05}_{0.11}$           &          $\textbf{49.69}_{0.00}$  \\ 
\midrule
LLaMA2-7B & n/a & 1.39 & 32.99  \\
GPT-3.5 & n/a & 1.19 & 33.32 \\ 
\midrule
\multicolumn{4}{c}{MIMIC-II} \\ \midrule 
THP & $6.63_{0.03}$ & $2.49_{0.29}$ & $85.05_{0.00}$
\\
SAHP & 
$2.48_{0.26}$ & $1.68_{0.06}$ & $84.11_{0.01}$
\\
ANHP & $1.85_{0.06}$ &$1.06_{0.06}$ & $83.14_{0.01}$ \\
\rowcolor[gray]{0.90} \ours & $\textbf{1.47}_{0.05}$ & $\textbf{1.06}_{0.01}$ & $\textbf{85.64}_{0.00}$   \\ \midrule
GPT-3.5 & n/a & 0.949 & 63.38 \\ \midrule
\multicolumn{4}{c}{Mooc} \\ \midrule 
Intensity Free$\dagger$                                  & $0.94_{0.03}$              & $0.31_{0.01}$              & $\textbf{40.00}_{0.01}$            \\
Neural Flows$\dagger$                                     & $0.43_{0.02}$                 & $0.47_{0.01}$                 & $30.00_{0.04}$     \\
THP+$\dagger$                                             & $0.13_{0.02}$              & $0.18_{0.01}$              & $38.00_{0.01}$          \\
Meta TPP$\dagger$                                         & $-0.72_{0.02}$              & $\textbf{0.16}_{0.01}$              &$ 36.00_{0.01}$           \\ \midrule
ANHP                                            & $-2.78_{0.02}$             & $0.20_{0.01}$              & $21.66_{0.01}$          \\
\rowcolor[gray]{0.90} \ours & $\textbf{-3.97}_{0.05}$    & $0.19_{0.01}$     & $31.49_{0.01}$ \\\midrule   
\multicolumn{4}{c}{Reddit} \\ \midrule 
Intensity Free$\dagger$  & $1.09_{0.04}$              & $0.18_{0.01}$              & $60.00_{0.01}$           \\
Neural Flows$\dagger$   & $1.30_{0.33}$              & $0.32_{0.04}$              & $60.00_{0.07}$           \\
THP+$\dagger$  & $1.20_{0.04}$              & $0.26_{0.01}$              & $60.00_{0.01}$           \\
Meta TPP$\dagger$   & $0.03_{0.04}$              & $\textbf{0.11}_{0.01}$              & $60.00_{0.01}$           \\\midrule
ANHP   & $0.05_{0.03}$              & $0.19_{0.00}$              & $\textbf{61.82}_{0.00}$  \\
\rowcolor[gray]{0.90} \ours & $\textbf{-0.25}_{0.03}$    & $0.18_{0.00}$     & $59.19_{0.00}$       \\\bottomrule  
\end{tabular}
\label{tab:tpp_main}
\end{table}

\paragraph{Downstream task 1: Temporal point process}
The experimental results on four real-world benchmark datasets are shown in \Cref{tab:tpp_main}. 
We compare with several state-of-the-art approaches, including SAHP \cite{sahp}, THP \cite{hawkes1}, Intensity-Free \cite{intensityfreetpp}, THP+ \cite{metatpp}, Meta TPP \cite{metatpp}, Neural Flows \cite{nf2021bilovs}, and ANHP \cite{anhp}.
On the Stack Overflow and MIMIC-II datasets, our approach achieves the lowest NLL and RMSE for event time prediction, and the highest accuracy for event type prediction, across all compared baselines. On the Mooc and Reddit datasets, our approach achieves the lowest NLL and competitive RMSE and accuracy scores. Since our method shares the same backbone encoder as ANHP, the ANHP baseline can be viewed as ablation of our method without pretext training. 
The results demonstrate the effectiveness of pretext training in predicting future events on real-world data.

\paragraph{Comparison to LLMs} Pre-trained large language models (LLMs) have been shown to be effective zero-shot forecasters \cite{llmtimeseries, llmtime2}. 
We test two representative LLMs on the next-event prediction task using Stack Overflow and MIMIC-II: the open-source LLaMA2-7B model~\cite{llama2} and the GPT-3.5-turbo-instruct model~\cite{gpt3}, which is a variant of GPT-3.5 fine-tuned for text completion. All predictions are made in a zero-shot setting as we do not fine-tune the LLMs on our datasets. Prompts to the model include a brief description of the next event prediction task, format requirements for inputs and outputs, a list of possible event type labels, and the history in an event sequence before the next event to predict. For each input, we sample 20 predictions from each model and filter out the samples that do not meet the output format requirements. 
Based on experiments on the validation set of Stack Overflow, we select the median of event time predictions among remaining samples (the smaller one of the middle two if the number of remaining samples is even) as the LLM's point estimate for the next event time to avoid extreme values. The event type corresponding to the prediction with the median event time is taken as LLaMA2's point estimate of event type, while GPT-3.5 uses the mode of event types among the samples.
We defer the exact prompts, output format requirements, and LLMs' hyper-parameters to the appendix.

\cref{tab:tpp_main} shows that LLMs can produce point estimates of the next event time with RMSE scores that are on par with temporal point process methods. There are some caveats in this comparison: the LLMs are pre-trained on much larger data sources and have much larger capacity, and they can produce only point estimates (no density estimation). Nevertheless, we think these are interesting findings that suggest that LLMs can be good zero-shot forecasters for event arrival times. The LLMs are less effective than temporal point process methods on event type prediction; this may be due to the lack of in-domain context.

\begin{table}[t]
\caption{AUC of sequence-level classification on the Mooc dataset.
Baseline refers to \ours\ without pretext training.
}
\centering
\begin{tabular}{lccc}
\toprule
    & Baseline & \ours \\
    \midrule
AUC &   0.7360       &   \textbf{0.7401}  \\
\bottomrule
\end{tabular}
\label{tab: seq classification}
\end{table}

\paragraph{Downstream task 2: Sequence-level classification}
The results of sequence-level classification are presented in \cref{tab: seq classification}. 
Each sequence in Mooc is associated with a course withdrawal meta-label (binary) and the overall label distribution is approximately balanced (57\% positive vs. 43\% negative).
We use the same transformer architecture trained from scratch without pretext training as our baseline.
Pretext training improves the AUC from 0.7360 to 0.7401.
The result suggests that our pretext training generalizes beyond temporal point processes and can also benefit sequence-level classification tasks.

\begin{figure}[t]
\centering
\includegraphics[width=0.49\columnwidth]{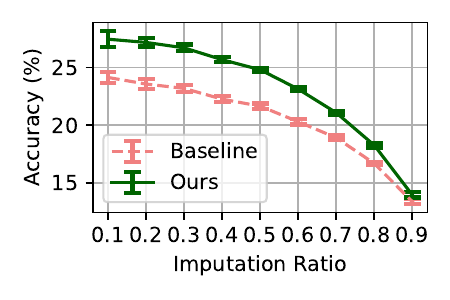}
\includegraphics[width=0.49\columnwidth]{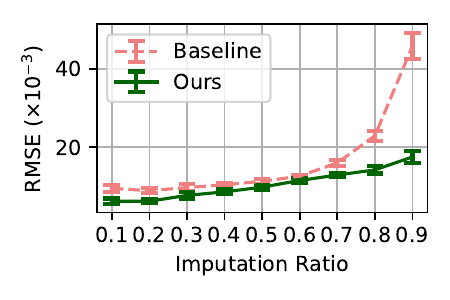}
\caption{Accuracy (higher is better) and RMSE (lower is better) of missing events imputation on the Mooc dataset. Baseline refers to \ours\ without pretext training. 
}
\label{tab: imputation}
\end{figure}

\paragraph{Downstream task 3: Missing events imputation} 
The results of missing events imputation are shown in \cref{tab: imputation} for the Mooc dataset. 
Our model is fine-tuned on the imputation task using a 50\% missing ratio and then tested on missing ratios from 10\% to 90\%; the baseline model is trained from scratch on the imputation task using a 50\% missing ratio and then tested on missing ratios from 10\% to 90\%.
We observe that pretext training improves imputation performance at all tested missing ratios.
Imputation performance decreases as the missing ratio increases, reflecting the increasing difficulty of the task given less information. 

\subsection{Ablation studies and additional experiments}\label{Sec: Ablation study}

In this section, we present ablation studies and additional experiments conducted on TPP with the Stack Overflow dataset.
Further experiments on model depth, model width, masking ratio, and masking strategy are deferred to \cref{appendix}. We also include an extension of our method to irregular time series.

\begin{table}[t]
    \centering
    \caption{Ablation study of the three pretext tasks on temporal point process prediction using the Stack Overflow dataset. ``Rec'', ``Cont'', and ``Align'' refer to masked reconstruction, contrastive learning, and alignment verification, respectively. } 
        \begin{tabular}{lcccccc}
            \toprule
            Methods & Rec & Cont & Align & NLL $\downarrow$  & RMSE $\downarrow$ & Acc $\uparrow$ \\
            \midrule
            Baseline     &  &  &  & 2.16 & 1.19 & 47.42 \\
            \midrule
            \multirow{7}{*}{\ours}    & \checkmark  &  &  & 1.86 & 1.13 & 49.68 \\
             &   & \checkmark &  & 1.84 & 1.16 & 49.80 \\
             &   &  & \checkmark & 1.86 & 1.36 & 47.96 \\
             & \checkmark  & \checkmark & & 1.83 & 1.13 & 49.28\\
             & \checkmark  &  & \checkmark & 1.84 & 1.23 & \textbf{49.81} \\
             &   & \checkmark & \checkmark & 1.82 & 1.07 & 49.75 \\
             & \checkmark  & \checkmark & \checkmark & \textbf{1.81} & \textbf{1.05} & 49.69 \\
            \bottomrule
        \end{tabular}
    \label{tab:ablation_modules}
\end{table}

\paragraph{Impact of different pretext tasks}
To evaluate the effectiveness of the different pretext tasks, \textit{i.e.}, masked reconstruction, contrastive learning, and alignment verification, we conduct systematic ablations as shown in \Cref{tab:ablation_modules}. 
The baseline, which does not employ any pretext training, serves as a reference point for comparison, achieving an NLL of 2.16, RMSE of 1.19, and accuracy of 47.42\%.
When applying each pretext task individually, consistent improvements are observed on the main NLL metric. 
Applying pairs of pretext tasks results in better NLL than applying the pretext tasks individually (e.g., using both contrastive learning and alignment verification produces a lower NLL than applying only contrastive learning or only alignment verification). The best pairing combines contrastive learning and alignment verification.
Finally, the overall best results are achieved when all three pretext tasks are employed together. The results of the ablation study indicate that the three proposed pretext tasks are complementary, and that an integrated strategy combining all three pretext tasks provides the most effective pre-training.

\begin{table}[t]
    \centering
    \caption{Few-shot performance of pretext training. We use a subset containing 25\%, 50\%, or 75\% of the training or pretext training dataset and report the NLL loss.
    ``PT'' and ``TR'' refer to the percentage of pretext training data or training data.
    }
    \begin{tabular}{lccccc}
        \toprule
        Methods & PT & TR & NLL $\downarrow$ & RMSE $\downarrow$ & Acc $ \uparrow$\\ \midrule
        Baseline & - & $100_{\%}$ & 2.16 & 1.19 & 47.42 \\ \midrule
        \multirow{7}{*}{\ours}& $100_{\%}$ & $25_{\%}$ & 1.88 & 1.17 & 49.72 \\
        & $100_{\%}$ & $50_{\%}$ & 1.86 & 1.03 & 49.75  \\
        & $100_{\%}$ & $75_{\%}$ &  1.87 & 1.24 & 49.72\\
        \cmidrule(l){2-6}
        & $25_{\%}$ & $100_{\%}$ & 1.92 & 1.11 & 49.14 \\
        & $50_{\%}$ & $100_{\%}$ & 1.87 & 1.14 & 49.90 \\
        & $75_{\%}$ & $100_{\%}$ & 1.85 & 1.10 & 49.71 \\
        \cmidrule(l){2-6}
        & 
        $100_{\%}$ & $100_{\%}$ & 1.81 & 1.05 & 49.69 \\ 
        \bottomrule
    \end{tabular}
    \label{tab:ablation_few_shots}
\end{table}

\paragraph{Few-shot pretext training}
We examine the ability of our pretext training method to generalize in few-shot settings by employing a subset of the dataset during both the pretext training and fine-tuning stages. Specifically, we utilize 25\%, 50\%, and 75\% of the entire training dataset and subsequently evaluate the model's performance on the complete test dataset. 
The experiment results are presented in \Cref{tab:ablation_few_shots}.
We find that fine-tuning our self-supervised representations on a fraction of the training dataset already outperforms the baseline network: our method using the full training data for pre-training and 25\% of the training data for fine-tuning outperforms the baseline trained on the full training data across all metrics. The result shows the few-shot learning potential of our method, making it valuable in scenarios with limited training resources.
When we vary the percentage of training data available for pre-training, we find that using even 25\% of the training data produces self-supervised representations that can be fine-tuned to outperform the baseline model across all metrics. As expected, the effectiveness of the self-supervised representations improves as more data is made available for pre-training.

\begin{table}[t]
    \centering
    \caption{Alternative time embedding methods. The results from using different time embeddings are shown in the upper section and their definitions are in the lower section.
    }
        \begin{tabular}{llccc}
        \toprule
        Methods & Time Embedding  & NLL $\downarrow$& RMSE $\downarrow$ & Acc $ \uparrow$\\\midrule
        \multirow{3}{*}{Baseline}  & Fixed \cite{anhp} & 2.16 & 1.19 & 47.42\\
         & Mercer \cite{mercer} & 2.13 & 1.16 & 46.50\\
          & mTAN \cite{mtan} & 2.19& 1.21 & 46.33\\
        \midrule
        \multirow{3}{*}{\ours}     & 
        Fixed \cite{anhp}  & 1.81 & 1.05 & 49.69 \\
          & Mercer \cite{mercer} & 1.93 & 1.18 & 48.77 \\
            & mTAN \cite{mtan}  & 2.01 & 1.16 & 48.97 \\
        \bottomrule
        \toprule
        \end{tabular}
    \resizebox{.72\columnwidth}{!}{%
        \begin{tabular}{ll}
            Fixed    & 
            \(
                \textbf{e}^t_i= \begin{cases}
                \cos \left( 10000^{-\frac{\lfloor{(i-1) / 2}\rfloor}{D_{time}}}  t_j  \right), & \text{ if } i \text{ is odd } \\ 
                \sin \left( 10000^{-\frac{\lfloor{(i-1) / 2}\rfloor}{D_{time}}}  t_j  \right), & \text{ if } i \text{ is even }
                \end{cases}\,.
            \)
            \\ \midrule
            Mercer &  
            \(
                \textbf{e}^t = 
            \left[\sqrt{c_1}, \ldots, \sqrt{c_{2 j}} \cos \left(\frac{j \pi t}{\omega}\right), \sqrt{c_{2 j+1}} \sin \left(\frac{j \pi t}{\omega}\right), \ldots\right]_{j\in[D_{time}]} \,.
            \)
            \\\midrule
            mTAN   & 
            \(
                \textbf{e}^t_i= \begin{cases}
                \mathbf{w}_1 t_1, & \text{ if } i = 1 \\ 
                \sin \left( \mathbf{w}_j  t_j  \right), & \text{ if } i \neq 1
                \end{cases}\,.
            \)
            \\ \bottomrule
        \end{tabular}
    }
    \label{tab:ablation_time_embedding}
\end{table}

\paragraph{Alternative time embeddings}
Different time embeddings have been found to be beneficial in the time series domain~\cite{mtan, mercer}. We tested three time embedding methods: fixed time embedding~\cite{anhp}, Mercer time embedding~\cite{mercer}, and a learnable time embedding~\cite{mtan}. The results are shown in \Cref{tab:ablation_time_embedding}. The three time embedding methods perform comparably on the baseline. With pretext training, the fixed time embedding outperforms the other methods across NLL, RMSE, and accuracy metrics.

\section{Conclusion}

Self-supervised learning enables the pre-training of generalizable representations, obviating the need to train models from scratch for different downstream tasks. In this paper, we presented three complementary self-supervised pretext tasks for discrete event sequences. In vision and language, a common paradigm is to pre-train on large data and then specialize to smaller domains, for example via fine-tuning or prompt-tuning. Our few-shot experiments suggest that this approach may be possible for event sequence data. Achieving this paradigm more broadly will require complementary progress in making large data available for pre-training event sequences. Interestingly, contemporary work in generating large-scale synthetic financial transactions \cite{altman2023} suggests that the research community is already making exciting progress in this direction.

\renewcommand{\refname}{\spacedlowsmallcaps{References}} 

\bibliographystyle{plainnat}

\bibliography{sample.bib} 


\newpage
\appendix

\section{Appendix}\label{appendix}

\subsection{Dataset details}\label{appendix:sec:datasets}

\textbf{StackOverflow} (SO)~\cite{snapnets}. 
It includes sequences of user awards within two years. 
StackOverflow is a question-answering website where users are awarded based on their proposed questions and their answers to questions proposed by others. 
This dataset contains a total of 6,633 sequences. 
There are $22$ types of events: Nice Question, Good Answer, Guru, Popular Question, Famous Question, Nice Answer, Good Question, Caucus, Notable Question, Necromancer, Promoter, Yearling, Revival, Enlightened, Great Answer, Populist, Great Question, Constituent, Announcer, Stellar Question, Booster and Publicist. 
The award time records when a user receives an award. 
With this dataset, we can learn which type of awards will be given to a user and when.

\textbf{Mooc}~\cite{kumar2019predicting}. 
It contains the interaction of students with an online course system. 
An interaction is an event and can be of various types ($97$ unique types), e.g. watching a video, solving a quiz etc.

\textbf{Reddit}~\cite{kumar2018community}. On this social network website, users submit posts to subreddits. In the dataset, the most active subreddits are selected, and posts from the most active users on those subreddits are recorded. Each sequence corresponds to a list of submissions a user makes. The data contains 984 unique subreddits that we use as classes in mark prediction.

\textbf{MIMIC-II}~\cite{DBLP:conf/embc/LeeSVCSM11}. 
The Multiparameter Intelligent Monitoring in Intensive Care (MIMIC-II) dataset is developed based on an electric medical record system. 
The dataset contains a total of 650 sequences, each of which corresponds to an anonymous patient’s clinical visits in a seven-year period. 
Each clinical event records the diagnosis result and the timestamp of that visit. 
The number of unique diagnosis results is $75$. 
According to the clinical history, a temporal point process is supposed to capture the dynamics of when a patient will visit doctors and the diagnosis result.

\subsection{Further ablation studies}

\begin{figure}[h]
    \centering
    \begin{minipage}{0.47\textwidth}
        \centering
        \includegraphics[width=\linewidth]{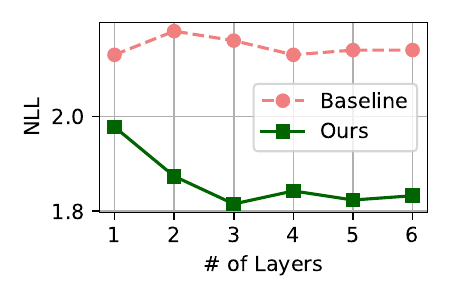}
        \caption{NLL of baseline (\ours\ w/o pretext training) and \ours\ on Stack Overflow versus the number of layers. }
        \label{tab:ablation_depth}
    \end{minipage}
    \hfill
    \begin{minipage}{.47\textwidth}
        \centering
        \includegraphics[width=\linewidth]{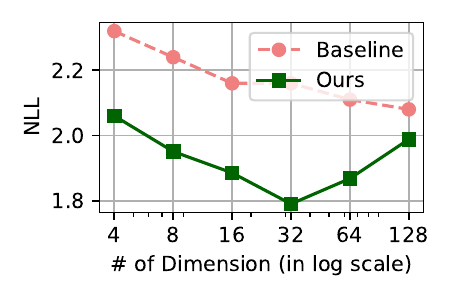}
        \caption{NLL of baseline (\ours\ w/o pretext training) and \ours\ on Stack Overflow versus the number of feature dimensions. }
        \label{tab:ablation_width}
    \end{minipage}%
\end{figure}

\noindent\textbf{How deep should the model be?} 
To understand the impact of the number of transformer blocks on our method, we vary the depth (number of layers) from 1 to 6 as shown in \Cref{tab:ablation_depth}. 
The baseline model exhibits relatively consistent but suboptimal performance across the range of layers, with NLL values ranging from 2.13 to 2.18. 
In contrast, our method (with pretext training strategies) consistently outperforms the baseline, demonstrating a progressive improvement in NLL from 1.98 with one layer to 1.81 with three layers. 
It underscores the effectiveness of our 
pretext training
in enhancing performance, suggesting that increasing the number of layers contributes to improved performance with the proposed pretext training methods.

\noindent\textbf{How wide should the model be?} 
To understand how wide the model should be (the feature dimension), we conduct experiments as shown in \Cref{tab:ablation_width}. 
We notice that, \ours\ consistently outperforms the baseline across all dimensionality settings, starting with a lower NLL of approximately 2.06 for 4 dimensions and achieving the lowest NLL of about 1.81 for 32 dimensions. 
And the best performance of \ours\ is achieved by setting the feature dimension to 32. 
After that, the model tends to suffer from the improvement of feature dimension as the NLL goes bigger, showing that \ours\ is able to learn reasonable representation with a limited number of features. 

\begin{table}
\centering
\caption{NLL of our method using different mask ratios of the reconstruction task.}
\begin{tabular}{lcccc}
\toprule
Methods               & Ratio & NLL $\downarrow$ & RMSE $\downarrow$ & Acc $ \uparrow$ \\ \midrule
Baseline              & -     & 2.16             & 1.19              & 47.42           \\ \midrule
\multirow{9}{*}{\ours} & 0.1   & 1.87             & 1.11              & 49.14           \\
                      & 0.2   & \textbf{1.80}             & 1.14              & \textbf{49.90}           \\
                      & \cellcolor[gray]{0.90} 0.3   & \cellcolor[gray]{0.90}1.81             & \cellcolor[gray]{0.90}1.05              & \cellcolor[gray]{0.90}49.69           \\
                      & 0.4   & 1.82             & \textbf{0.97}              & 49.74           \\
                      & 0.5   & 1.85             & 0.98              & 49.65           \\
                      & 0.6   & 1.84             & 1.12              & 49.83           \\
                      & 0.7   & 1.82             & 1.04              & 49.84           \\
                      & 0.8   & 1.84             & 1.10              & 49.81           \\
                      & 0.9   &       1.87           &        1.42           &   49.73             \\
                      \bottomrule
\end{tabular}
    \label{tab:ablation_reconstruction_ratio}
\end{table}

\noindent\textbf{How much masking is needed?} 
To determine the best configuration of masking data, we conducted a series of experiments with different masking ratios as shown in \cref{tab:ablation_reconstruction_ratio}. 
We vary mask ratios from 0.1 to 0.9. 
Notably, as the mask ratio increases, the NLL score tends to decrease, reaching its lowest value of 1.80 at a mask ratio of 0.2. 
This demonstrates the effectiveness of introducing a moderate level of reconstruction masking for improved model performance. 
Additionally, the RMSE metric follows a similar trend, with the lowest value of 0.97 achieved at a mask ratio of 0.4. 
The highest accuracy score of 49.90\% is attained at a mask ratio of 0.2, highlighting the optimal balance between masking and reconstruction for our method.

Overall, considering that the target of fine-tuning is minimizing NLL, as suggested by \cref{tab:ablation_reconstruction_ratio}, a moderate mask ratio, such as 0.3, yields the most favorable trade-off, resulting in low NLL score, competitive RMSE, and accuracy values.

\begin{table}
    \centering
    \caption{NLL of our method using random masking or density preserving masking.}
        \begin{tabular}{lccccc}
        \toprule
        Methods & Random & Density  & NLL $ \downarrow$              & RMSE $\downarrow$ & Acc $ \uparrow$ \\
        \midrule
        Baseline  &  &   & 2.16 & 1.19 & 47.42 \\
        \midrule
        \multirow{2}{*}{\ours}     &\checkmark&     &  1.83 & \textbf{1.05} & \textbf{49.82} \\
          &\cellcolor[gray]{0.90}  &\cellcolor[gray]{0.90}  \checkmark &\cellcolor[gray]{0.90} \textbf{1.81} & \cellcolor[gray]{0.90}\textbf{1.05}           &\cellcolor[gray]{0.90} 49.69 \\
        \bottomrule
        \end{tabular}
\label{tab:ablation_constant_random_masking}
\end{table}

\noindent\textbf{Random masking or density-preserving masking?} 
\citet{primenet} proposed the constant duration masking strategy for irregular time series data, taking the asynchronous nature of time gaps between observation into account, which are similar to the interarrival times between events in event sequence data. 
To better understand how this strategy affects our methods, we report the results using constant time masking (density-preserving masking) and random masking in \Cref{tab:ablation_constant_random_masking}. 
We find that both masking strategies perform comparably in our context, with density-preserving masking achieving slightly better results on the main NLL metric.

\subsection{Generalization to irregular time series}

To show the generalization ability of our method, we also evaluate our proposed method on two representative irregular time series processing tasks, classification and interpolation. 

\paragraph{Baselines, datasets, and evaluation protocols}
We compare our methods with PrimeNet~\cite{primenet}, a pretraining method in irregular time series analysis 
and 2 irregular time-series methods (ODE-RNN~\cite{ODERNN} and mTAND~\cite{mtan}). 
Following \citet{primenet}, we consider interpolation and classification as downstream tasks on PhysioNet~\cite{silva2012predicting} and Human Activity~\cite{DBLP:conf/ami/KaluzaMDLG10} datasets. 

PhysioNet Challenge 2012~\cite{silva2012predicting}  is a multivariate time series dataset consisting of 37 physiological variables, respectively, extracted from intensive care unit (ICU) records. 
Each record contains sparse and irregularly spaced measurements from the first 48 hours after admission to the ICU. 
We predict in-hospital mortality (binary classification) from this data. PhysioNet has 13.8\% positive data. Human Activity~\cite{DBLP:conf/ami/KaluzaMDLG10} dataset has 3-D positions of the waist, chest, and ankles from 5 individuals performing activities including walking, sitting, lying, standing, etc.

We use RMSE as the evaluation metric for interpolation task on both datasets. The evaluation metrics for classification task are area under the ROC curve (AUC) and accuracy on PhysioNet and Human Activity respectively. For interpolation, models infer a continuous missing segment of 10\% and 50\% values with the observation of the remaining data. 
The model is fine-tuned with cross entropy loss for classification and RMSE for interpolation. 
For all the experiments, we follow the recommended hyperparameters and use the same architecture with PrimeNet~\cite{primenet}, i.e., a 2-layer transformer-based model with the hidden size of 128 and 1 head. 
Specifically, the classification head is a 3-layer MLP with ReLU activation ($128\rightarrow 300 \rightarrow n_{class}$), while for the interpolation head, we use a 5-layer MLP with the ReLU activation ($128\rightarrow 512 \rightarrow 512 \rightarrow 512 \rightarrow 512 \rightarrow 1$). 
For a fair comparison, we add our proposed masked reconstruction, contrastive learning, and alignment verification strategies to PrimeNet as \ours.

\begin{table}
\centering
\caption{Classification (AUC or Accuracy) and interpolation (RMSE) results on PhysioNet and Human Activity. Higher (for classification) / Lower (for interpolation) is better and best in \textbf{bold}. ``*'' refers to results reproduced by us using public official codes.
}
\begin{tabular}{lcccc}
\toprule
\multirow{2}{*}{Methods}& \multicolumn{2}{c}{PhysioNet} & \multicolumn{2}{c}{Human Activity}  \\ 
\cmidrule(l){2-3}  \cmidrule(l){4-5}
 & Cls. (AUC) $\uparrow$ & Inp. ($\times10^{-2}$) $\downarrow$ & Cls. (Acc.) $\uparrow$ & Inp. ($\times10^{-2}$) $\downarrow$\\
\midrule
ODE-RNN &  0.694 & 11.91 & 0.885 & 26.69  \\
mTAND &  0.837 & 6.89 & 0.918 & 20.46\\\midrule
PrimeNet &  0.842 & 4.78* & 0.913 & 14.30 \\
\rowcolor[gray]{0.90} \ours & \textbf{0.849} & \textbf{4.55} & \textbf{0.920} & \textbf{6.59} \\
\bottomrule
\end{tabular}
\label{tab:irr_table}
\end{table}

\paragraph{Classification}
The results are shown in \Cref{tab:irr_table}.
On two different benchmarks, \ours\ consistently stands out as the top-performing method. 
It achieves the highest AUC and accuracy scores among all methods, indicating its superior ability to correctly classify irregular time series data. Specifically, 
the AUC on PhysioNet and the accuracy on Human Activity of ours are 0.849 and 0.920 respectively, compared to 0.842 and 0.918 of the best baseline models.
This result highlights our framework as a promising approach for time series classification tasks.

\paragraph{Interpolation}
We further evaluate our approach on the missing value interpolation task with results shown in \Cref{tab:irr_table}.
The interpolation task assesses the models' ability to predict missing data in a time series accurately given the observed part. Lower RMSE values indicate better interpolation accuracy. 
Our proposed method once again demonstrates its superiority, achieving the lowest RMSE value among all methods. 
Achieving RMSEs of 0.0455 and 0.0659 on Physionet and Human Activity respectively, \ours\ outperforms the baseline models with best RMSEs of 0.0478 and 0.1430.
This outcome underscores \ours' superior accuracy in interpolating missing data points for time series. 

\subsection{Large Language Model experiment details}
Our prompts for zero-shot next event prediction are adapted from the prompt used in the code base of \textsc{LLMTime}~\cite{llmtimeseries}, which applied large language models to zero-shot time series forecasting. The prompt we used for Stack Overflow~\cite{snapnets} dataset is presented in Prompt~\ref{prompt:so}.
\lstset{
  basicstyle=\ttfamily,
  columns=fullflexible,
  frame=single,
  breaklines=true,
  breakindent=0pt
}
\begin{lstlisting}[upquote=true, caption={LLM Prompt for Zero-shot Next Event Prediction on Stack Overflow}, label={prompt:so}, captionpos=b]
You are a helpful assistant that performs event sequence predictions. The user will provide a sequence of events and you will predict the next event. The sequence is represented by a sequence of tuples in parenthesis such that the first element is a string of event type and second element is a float point number of time since the last event. Please predict the next event in the same format of event in the input without producing any additional text. The input sequence may contain only one event or be empty. In the case of empty input sequence, just predict the first event. Do not say anything like 'the next event in the sequence are', just return the next event in a tuple. The types of event you are allowed to predict as the first element in your response must be in the following list: Guru, Nice Question, Necromancer, Caucus, Announcer, Nice Answer, Booster, Great Answer, Revival, Stellar Question, Famous Question, Enlightened, Notable Question, Popular Question, Populist, Great Question, Good Answer, Constituent, Yearling, Good Question, Publicist, Promoter. Sequence: \n 
\end{lstlisting}

Following a common practice in TPP research work, we used an de-identified version of the MIMIC-II dataset~\cite{anhp}, making it challenging to map the pre-processed labels in integers back to plain text. Therefore, we directly use the integer labels in the inputs and prompts for MIMIC-II dataset and the prompt is presented in Prompt~\ref{prompt:mimic}.
\lstset{
  basicstyle=\ttfamily,
  columns=fullflexible,
  frame=single,
  breaklines=true,
  breakindent=0pt
}
\begin{lstlisting}[frame=single,float=!htb,upquote=true, caption={LLM Prompt for Zero-shot Next Event Prediction on MIMIC-II},label={prompt:mimic}, captionpos=b]
You are a helpful assistant that performs event sequence predictions. The user will provide a sequence of events and you will predict the next event. The sequence is represented by a sequence of tuples in parenthesis such that the first element is an integer indicating event type and second element is a float point number of time since the last event. Please predict the next event in the same format of event in the input without producing any additional text. The input sequence may contain only one event or be empty. In the case of empty input sequence, just predict the first event. Do not say anything like 'the next event in the sequence are', just return the next event in a tuple. TThe integer label of event types you are allowed to predict as the first element in your response must be in the following list: [0, 1, 2, 3, 4, 5, 6, 7, 8, 9, 10, 11, 12, 13, 14, 15, 16, 17, 18, 19, 20, 21, 22, 23, 24, 25, 26, 27, 28, 29, 30, 31, 32, 33, 34, 35, 36, 37, 38, 39, 40, 41, 42, 43, 44, 45, 46, 47, 48, 49, 50, 51, 52, 53, 54, 55, 56, 57, 58, 59, 60, 61, 62, 63, 64, 65, 66, 67, 68, 69, 70, 71, 72, 73, 74]. Sequence: \n  
\end{lstlisting}

The prompt is followed by a sequence of inputs in the format of \verb|(<EVENT TYPE>|, \verb|<TIME SINCE LAST EVENT>)| separated by comma and white space. For LLaMA2 model, we append additional prompt \verb|\n the next is ==>| to the input. 
For each input, we generate 20 samples of next event predictions from LLM and take the following steps to post-process the output string and filter out invalid predictions: 1) We remove all white space, including \verb|\n| and capitalize all the letters in each output; 2) the leading left parenthesis and ending right parenthesis are removed from the output if they exist; 2) for an output after Step 1 and 2 to be a valid prediction, we require the string to contain exactly one comma, the sub-string after the comma must be convertible to a floating point number, and the sub-string before the comma must be one of the allowed event type labels (with letters capitalized and white space removed). We use the 50 sequences from the validation (dev) set of Stack Overflow to tune the temperature of the LLMs which controls the stochasticity of sampling. The temperatures are 0.8 and 2 for GPT3.5 and LLaMA2 respectively.

\end{document}